\documentclass[letterpaper, 10 pt, conference]{cls/ieeeconf}

\IEEEoverridecommandlockouts
\usepackage{balance}
\usepackage{silence}
\IEEEoverridecommandlockouts

\usepackage{cite}
\usepackage{graphicx}
\usepackage{picinpar}
\usepackage{stfloats}
\usepackage{url}
\usepackage{flushend}
\usepackage[latin1]{inputenc}
\usepackage{colortbl}
\usepackage{soul}
\usepackage{multirow}
\usepackage{pifont}
\usepackage{alltt}
\usepackage{enumerate}
\usepackage{siunitx}
\usepackage{pbox}
\usepackage{amsmath,amssymb,amsfonts}
\usepackage{amsthm}
\usepackage{algorithmic}
\usepackage{algorithm}
\usepackage{bm}
\usepackage{graphbox}
\usepackage{textcomp}
\usepackage[dvipsnames]{xcolor}
\usepackage[export]{adjustbox}
\usepackage{color}
\usepackage{subcaption}
\usepackage{makecell}
\usepackage[most]{tcolorbox}  
\usepackage{paralist}
\usepackage[frozencache,cachedir=_minted]{minted}
\tcbuselibrary{hooks, minted}

\newtcblisting[auto counter,number within=section]{mycodebox}[3][]{
  colback=gray!10,          
  colframe=black,           
  coltitle=white,           
  fonttitle=\bfseries,      
  title={\scriptsize \thetcbcounter: {#3}}, 
  title style={height=5pt, before skip=0pt, after skip=0pt},
  minted options app={mathescape, fontsize=\scriptsize, numbersep=5pt, breaklines, breaksymbolleft={}},  
  minted language={#2},
  #1,
  listing only,
  sharp corners,
  boxrule=1pt,          
  left=0pt, right=0pt, top=0pt, bottom=0pt  
}

\newcommand{\AP}[0]{\calA\calP}

\definecolor{LightCyan}{rgb}{0.8,0.8,1.0}
\definecolor{LightRed}{rgb}{1.0,0.8,0.8}
\definecolor{LightGreen}{rgb}{0.8,1.0,0.8}
\definecolor{LightYellow}{rgb}{1.0,1.0,0.8}

\theoremstyle{definition}
\newtheorem{definition}{Definition}
\newtheorem*{problem}{Problem}

\theoremstyle{remark}

\ActivateWarningFilters[balance]

\makeatletter
\let\NAT@parse\undefined
\makeatother
\usepackage[hidelinks]{hyperref}

\newcommand{\calA}{{\cal A}}
\newcommand{\calB}{{\cal B}}
\newcommand{\calC}{{\cal C}}

\newcommand{\calE}{{\cal E}}
\newcommand{\calF}{{\cal F}}
\newcommand{\calG}{{\cal G}}

\newcommand{\calP}{{\cal P}}
\newcommand{\calQ}{{\cal Q}}

\newcommand{\calV}{{\cal V}}

\newcommand{\calX}{{\cal X}}

\newcommand{\calZ}{{\cal Z}}

\newcommand{\bfv}{\mathbf{v}}

\newcommand{\bfx}{\mathbf{x}}
\newcommand{\bfy}{\mathbf{y}}

\newcommand{\bfF}{\mathbf{F}}
\newcommand{\bfG}{\mathbf{G}}

\newcommand{\bfU}{\mathbf{U}}

\newcommand{\bfX}{\mathbf{X}}

\newcommand{\bbN}{\mathbb{N}}

\newcommand{\bbR}{\mathbb{R}}

\title{\LARGE\bf LTLCodeGen: Code Generation of Syntactically Correct Temporal Logic for Robot Task Planning}

\author{Behrad Rabiei$^{*}$ \and Mahesh Kumar A.R.$^{*}$ \and Zhirui Dai \and Surya L.S.R. Pilla \and Qiyue Dong \and  Nikolay Atanasov
\thanks{$^{*}$Equal contribution.}%
\thanks{We gratefully acknowledge support from ARL DCIST CRA W911NF17-2-0181 and ONR N00014-23-1-2353.}%
\thanks{The authors are with the Contextual Robotics Institute, University of California San Diego, La Jolla, CA 92093, USA, e-mails: {\tt\small \{brabiei,\allowbreak ar223,\allowbreak zhdai,\allowbreak slpilla,\allowbreak q4dong,\allowbreak natanasov\}@ucsd.edu}.}%
}

\begin{document}
\maketitle


\begin{abstract}
This paper focuses on planning robot navigation tasks from natural language specifications. We develop a modular approach, where a large language model (LLM) translates the natural language instructions into a linear temporal logic (LTL) formula with propositions defined by object classes in a semantic occupancy map. The LTL formula and the semantic occupancy map are provided to a motion planning algorithm to generate a collision-free robot path that satisfies the natural language instructions. Our main contribution is LTLCodeGen, a method to translate natural language to syntactically correct LTL using code generation. We demonstrate the complete task planning method in real-world experiments involving human speech to provide navigation instructions to a mobile robot. We also thoroughly evaluate our approach in simulated and real-world experiments in comparison to end-to-end LLM task planning and state-of-the-art LLM-to-LTL translation methods.
\end{abstract}


\section{Introduction}\label{sec:intro}
\bstctlcite{IEEEexample:BSTcontrol}

The ability to understand and execute tasks specified in natural language is an important aspect of enabling autonomous robot assistants in our daily lives. A fundamental challenge is to translate high-level natural-language task descriptions into low-level, executable robot actions. Large Language Models (LLMs), such as GPT-4 \cite{openai2024gpt4technicalreport}, LLaMA-2 \cite{llama22023}, and DeepSeek-R1 \cite{deepseek2025}, have demonstrated remarkable natural language proficiency and language-based task reasoning~\cite{saycan2023,roco2024} and planning capabilities~\cite{gtb2024}. LLMs play different roles in the planning of robot task executions for natural language specifications \cite{dai2024optimal,saycan2023,codepolicy2023,chatenv2023,progprompt2023,smartllm2024,spine2024,roco2024,pddlllm2024}. An LLM can be a context extractor, which connects natural language concepts to real-world objects and locations, a process known as symbol grounding \cite{symbolgrounding2011}. LLMs are also used as translators, for example, to convert natural language instructions into linear temporal logic (LTL) formulas \cite{dai2024optimal}, signal temporal logic (STL) \cite{roco2024,pddlllm2024}, planning domain definition language (PDDL) \cite{pddlllm2024}, or executable code \cite{codepolicy2023,progprompt2023,demo2code2023}. An LLM can also operate as a task scheduler, which decomposes the task into sub-tasks, infers the order and the dependencies among the sub-tasks, and distributes them to different robots \cite{roco2024,smartllm2024}. LLMs also may cooperate with other multi-modality models to better fuse environment feedback into task planning and execution \cite{chatenv2023,spine2024}. However, recent works \cite{gtb2024,kambhampati2024llmscantplanhelp,valmeekam2023planningabilitieslargelanguage} have shown that LLMs may not always handle the whole process of converting natural language instructions into executable robot actions robustly without careful design of different components including symbol grounding, task decomposition, sub-task arrangement, formal logic translation, path generation, execution monitoring, and closed-loop feedback.

\begin{figure}[t]
    \centering
    \includegraphics[width=\linewidth]{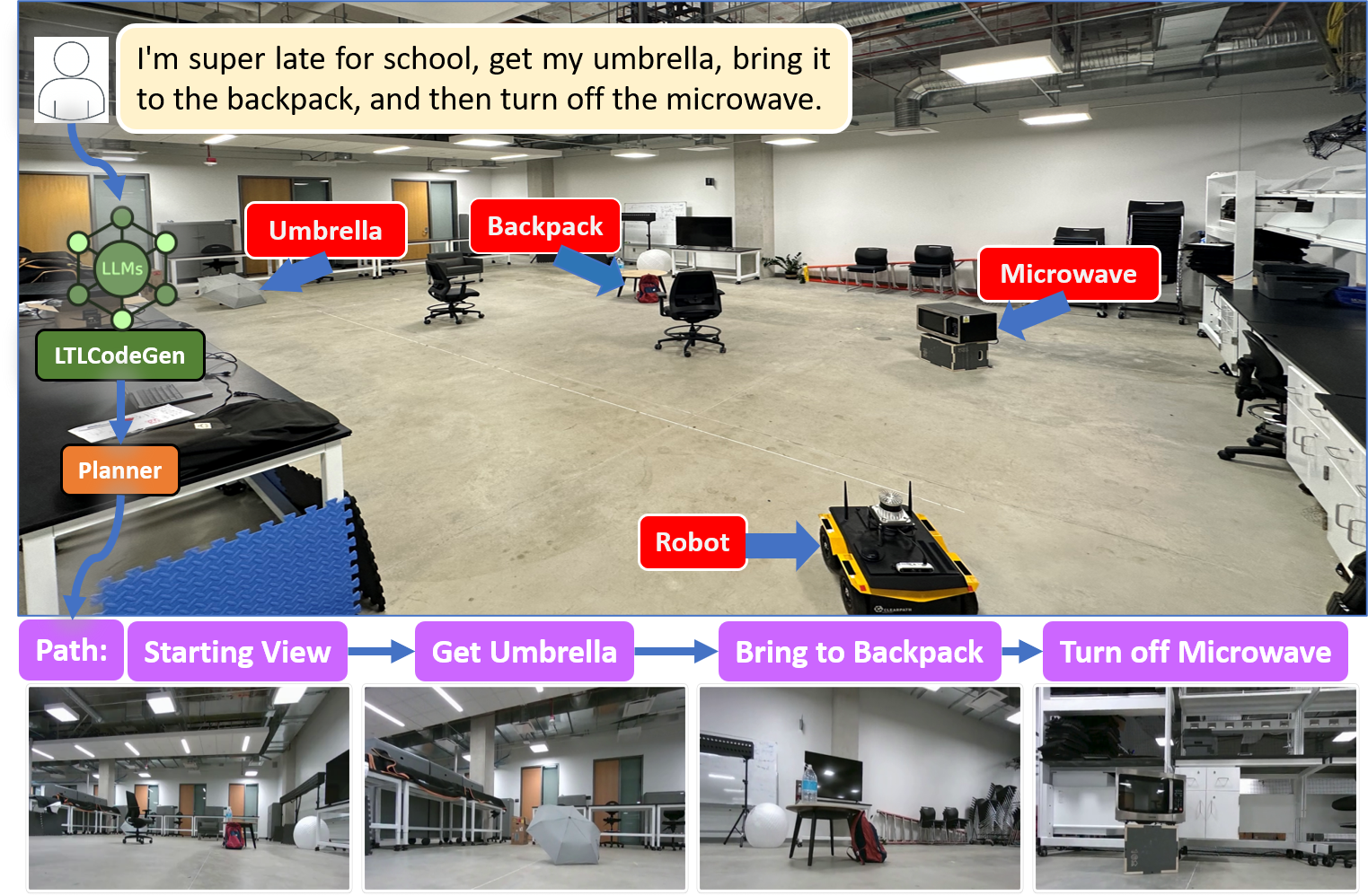}
    \caption{Our method is able to interpret natural language instructions for a robot navigation task, translate them into an LTL formula free of syntax errors, and plan a robot path that satisfies the instructions. The figure shows an example of a Jackal robot fetching an umbrella, bringing it to a backpack, and finally navigating to a microwave oven.}
    \label{fig:teaser}
\end{figure}


Many works have explored an end-to-end task planning approach where an LLM converts natural language instructions directly into robot actions \cite{innermonologue2022,chatenv2023,codepolicy2023,instruct2act2023,progprompt2023,roco2024}. In this paper, we consider an alternative modular task planning approach where an LLM converts natural language instructions into a task model first and then composes it with a robot model and uses task and motion planning to generate robot actions. To ground the concepts in natural language instructions, our approach relies on a semantic occupancy map \cite{ssmi2023} constructed from range observations and semantic segmentation observations \cite{yolo2023}. The map captures the free and occupied space as well as the categories of objects around the robot, allowing a connection to the objects that the natural language instructions may be referring to. We introduce a novel code generation approach, LTLCodeGen, to convert the natural language instructions into an LTL formula. The LTL formula and the semantic occupancy map are provided to a motion planning algorithm to generate a collision-free path that satisfies the task requirements. An example of our task planning approach is shown in Fig.~\ref{fig:teaser}.



The contributions of this paper are summarized as follows.
\begin{enumerate}
    \item We propose LTLCodeGen, a method to translate natural language to syntactically correct linear temporal logic formulas using code generation. 
    
    \item We develop a modular task planning approach that converts natural language instructions to LTL over semantic occupancy map and uses motion planning to generate robot paths that satisfy the task requirements.
    
    
    \item We compare LTLCodeGen with NL2LTL~\cite{dai2024optimal} and LLMs fine-tuned for LTL translation \cite{dataefficientltl2023} on both human-written and synthetic datasets. We also compare our task planning approach with end-to-end LLM-based task planning on the GameTraversalBenchmark (GTB) \cite{gtb2024}. Additionally, we conduct real-world robot experiments with tasks of different difficulty levels, and perform ablation studies to analyze the importance of different components in LTLCodeGen. 
\end{enumerate}

 \section{Related work}\label{sec:related}


The components of natural language task planning include a) translating natural language instructions into a machine-interpretable format; b) constructing a scene representation with rich information to support instruction grounding and task planning; c) planning the task based on the translated instruction and the scene representation.

\subsection{Natural Language Instruction Translation}
Regardless of whether LLMs understand human language, their ability to transform text input to different text outputs based on specific requirements is evident \cite{climbing2020}. Since LLMs are trained on internet-scale text data, the common knowledge stored in the model is helpful for inferring the human intent embedded in the natural language instructions. Converting instructions into machine-interpretable format involves two challenges.
One is to associate concepts, e.g. objects, in the instructions, to physical locations in space, which is known as symbol grounding. Whether LLMs solve symbol grounding well remains unclear \cite{climbing2020,gubelmann2024} due to the debate on whether LLMs understand human language. The other challenge is to translate the symbol-grounded instruction to a machine-interpretable format that must correctly encode both the task objectives and constraints.

Recent works use different task representation formats, including LTL \cite{dataefficientltl2023,dai2024optimal}, STL~\cite{autotamp2024,nl2tl2024}, PDDL~\cite{pddlllm2024}, code~\cite{codepolicy2023,progprompt2023,demo2code2023,smartllm2024}, or others (e.g. waypoints~\cite{roco2024}, action scores~\cite{saycan2023}, API calls~\cite{spine2024}). 
Because LLMs are usually trained with a large corpus of code examples, they excel in writing high-quality code that calls predefined APIs based on the instruction, as shown in CaP~\cite{codepolicy2023}, ProgPrompt~\cite{progprompt2023}, and Demo2Code~\cite{demo2code2023}. However, it is still a challenge to generate code that provides an optimal and semantically correct plan for direct task execution. Other works use LTL \cite{dai2024optimal} or STL \cite{autotamp2024}, whose alignment with the user intent can be examined, to provide a task constraint for motion planning algorithms \cite{tulip2011,vasile2013sampling,optimaltl2016}. However, LLMs are not trained with various temporal logic examples and, even with carefully designed examples in the prompt, LLMs still generate syntactically incorrect LTL formulas occasionally.
Some works \cite{dataefficientltl2023,nl2tl2024} propose to fine-tune LLMs for translating temporal logic. However, it is still difficult to scale up as the LTL formula becomes more complicated. 
To obtain both the syntax robustness of code generation and the post-processing convenience of temporal logic, we develop LTLCodeGen, which uses LLMs to generate code that outputs LTL formulas. LTLCodeGen makes the whole system more robust to handle more complicated tasks and enables optimal planning for the instruction \cite{optimaltl2016,dai2024optimal}.

\subsection{Scene Representation}
A semantically annotated scene representation is essential for accurate instruction interpretation. It must support symbol grounding, collision checking, and task progress feedback to ensure safe and correct execution.
Some works aim to build such representations, such as a semantic occupancy octree~\cite{ssmi2023}, or a scene graph \cite{armeni20193d,rosinol20203d,kimera,hughes2022hydra,amiri2022reasoning}.
However, LLMs have a limited context window size, so the scene should be described to the LLM in a compact way. Existing works provide descriptions of involved objects only \cite{codepolicy2023,autotamp2024,smartllm2024} or compress the scene in a concise format and then rely on the LLM to extract items related to the instruction \cite{dai2024optimal}. Other works also consider obtaining scene information in different modalities \cite{socratic2022,chatenv2023}. 

\subsection{Task and Motion Planning}
Prior works focus on different aspects of task and motion planning for natural language instruction. For improved feasibility, SayCan~\cite{saycan2023} fuses robot-state-based action affordances into the LLM planning procedure. InnerMonologue \cite{innermonologue2022} continually injects new sensor observations as feedback into the LLM planner to correct the execution online.
Some works focus on multi-robot collaboration using a single LLM to reason about task decomposition and allocation \cite{smartllm2024} or a dialogue between multiple LLMs \cite{roco2024}. 
In contrast, our work focuses on generating an optimal path satisfying the LTL formula that correctly reflects the instruction.


\section{Problem Statement}\label{sec:prosta}

Enabling mobile robots to navigate following human instructions in natural language is essential for effective human-robot interaction. We consider a mobile robot with state $x_t \in \calX \subseteq \bbR^n$ at time $t$ equipped with an RGBD sensor. The robot first needs to construct a map of its environment, which stores information needed to understand the semantics of a human instruction. To ensure safe navigation, the map should also capture the presence of obstacles correctly.

\begin{definition} \label{def:semantic_occupancy_map}
A \emph{semantic occupancy map} is a function $m: \calX \to \calC$, which associates a robot state $\bfx \in \calX$ to a semantic category $m(\bfx) \in \calC$ present at state $\bfx$. The set $\calC$ is a set of semantic categories, including object classes, a \textsc{free} class that represents unoccupied free space, a \textsc{null} class that represents occupied space with unknown semantic category, and an \textsc{unknown} class for unobserved space.
\end{definition}

\begin{figure}[t]
    \centering
    \includegraphics[width=\linewidth,trim={440pt 480pt 350pt 480pt},clip]{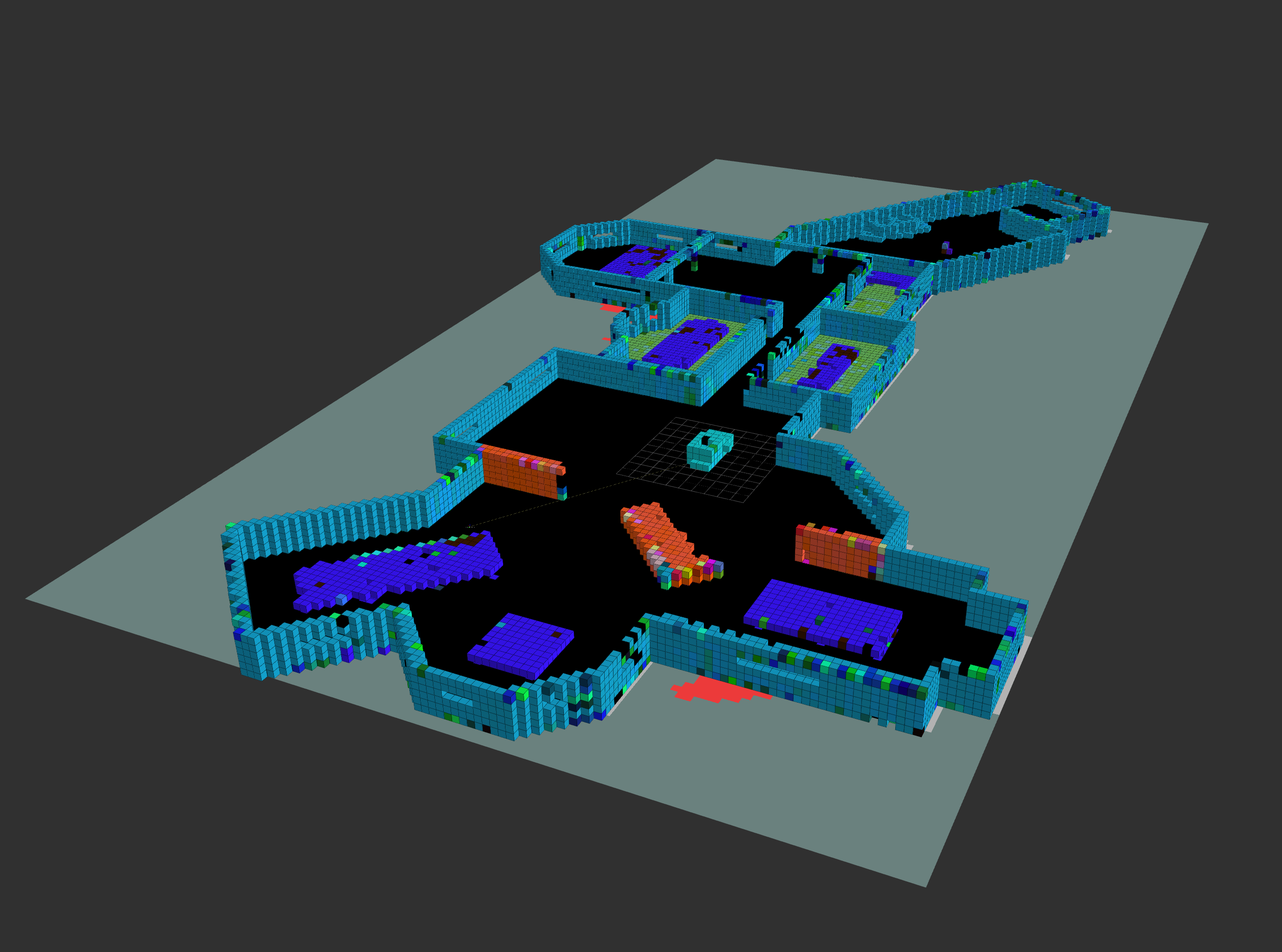}
    \caption{Semantic occupancy map of an office environment. Each voxel encodes both occupancy and semantic classification, represented by distinct colors: \textcolor{cyan}{cyan} indicates walls, \textcolor{orange}{orange} and \textcolor{blue}{blue} denote office tables, \textcolor{SkyBlue}{sky blue} represents a cart, and \textcolor{LimeGreen}{lime green} corresponds to carpeted floors.}
    \label{fig:octree}
\end{figure}

Fig.~\ref{fig:octree} shows an example of a semantic occupancy map. For example, given a semantic occupancy map of a house, the robot may be asked to tidy up the dining table and place the dishes in a dishwasher. Using the map, the robot can connect the instructions to the physical locations of the table and dishwasher and plan its motion to navigate to these locations safely and follow the sequence required by the instructions.

To construct a connection between the natural language instructions and the semantic occupancy map, we define boolean propositions that capture the task requirements.

\begin{definition} \label{def:atomic_proposition}
An \emph{atomic proposition} is a boolean function $p_c(\bfx)$ that evaluates true when the robot state $\bfx$ is sufficiently close to a class $c \in \calC$ in a semantic occupancy map $m$:
\begin{equation} \label{eq:atomic_proposition}
    p_c(\bfx) =
    \begin{cases}
        \text{true}, & \exists \bfy \in \calX, m(\bfy)=c, \left\|\bfx-\bfy\right\| \le r_c\\
        \text{false}, & \text{otherwise},
    \end{cases}
\end{equation}
where $r_c$ is a distance threshold associated with class $c$. Denote the set of atomic propositions by $\calA\calP = \{ p_c | c \in \calC\}$.
\end{definition}

To capture which atomic propositions are satisfied at different robot states, we define a label map. 

\begin{definition} \label{def:label_map}
    Given a semantic occupancy map $m$ and atomic propositions $\calA\calP$, a \emph{label map} $l: \calX$ $\to 2^{\calA\calP}$ associates a robot state $\bfx$ to the set of atomic propositions that evaluate true at $\bfx$, i.e., $p_c(\bfx) = \text{true}$ for all $p_c \in l(\bfx) \subseteq \calA\calP$.
\end{definition}



The atomic propositions satisfied along a path $\bfx_{1:T} := \bfx_1, \bfx_2, \ldots,\bfx_T$ are obtained from the label map as $l(\bfx_{1:T}) := l(\bfx_1), l(\bfx_2), \ldots l(\bfx_T)$. The sequence $l(\bfx_{1:T})$ is called a \emph{word} and it encodes the task requirements satisfied along the robot path $\bfx_{1:T}$.  Finally, we use the notation $l(\bfx_{1:T}) \models \mu$ to indicate that all requirements of a task $\mu$ are satisfied by a word $l(\bfx_{1:T})$. We define the motion planning problem associated with a natural language task $\mu$ as follows.


\begin{problem}
    Given a semantic occupancy map $m$, a natural language navigation task $\mu$ defined in terms of semantic classes $\calC$, a cost function $d: \calX \times \calX \to \bbR_{>0}$, and an initial robot state $\bfx_1 \in \calX$, plan a path $\bfx_{1:T}$ that satisfies $\mu$ with minimum cost:
    \begin{equation}
    \begin{aligned}
        \min_{T \in \bbN, \bfx_{1:T}} & \sum_{t=1}^{T-1} d(\bfx_t, \bfx_{t+1}) \\
        \text{s.t.} \quad l(\bfx_{1:T}) \models \mu; & \quad m(\bfx_t) = \textsc{free}, \;\; t = 1, \dots, T.
    \end{aligned}
    \end{equation}
\end{problem}

A key aspect of solving the problem above is to ground the task specification $\mu$ to states in the map $m$ with associated semantic classes and atomic propositions and to encode the dependencies among the atomic propositions according to $\mu$. In the next section, we propose an LLM code generation technique to generate LTL from $\mu$ using the atomic propositions defined by the semantic occupancy map $m$.

\section{Robot Task Planning via LTLCodeGen}
\label{sec:overview}

\begin{figure*}[t]
    \centering
    \includegraphics[width=\linewidth]{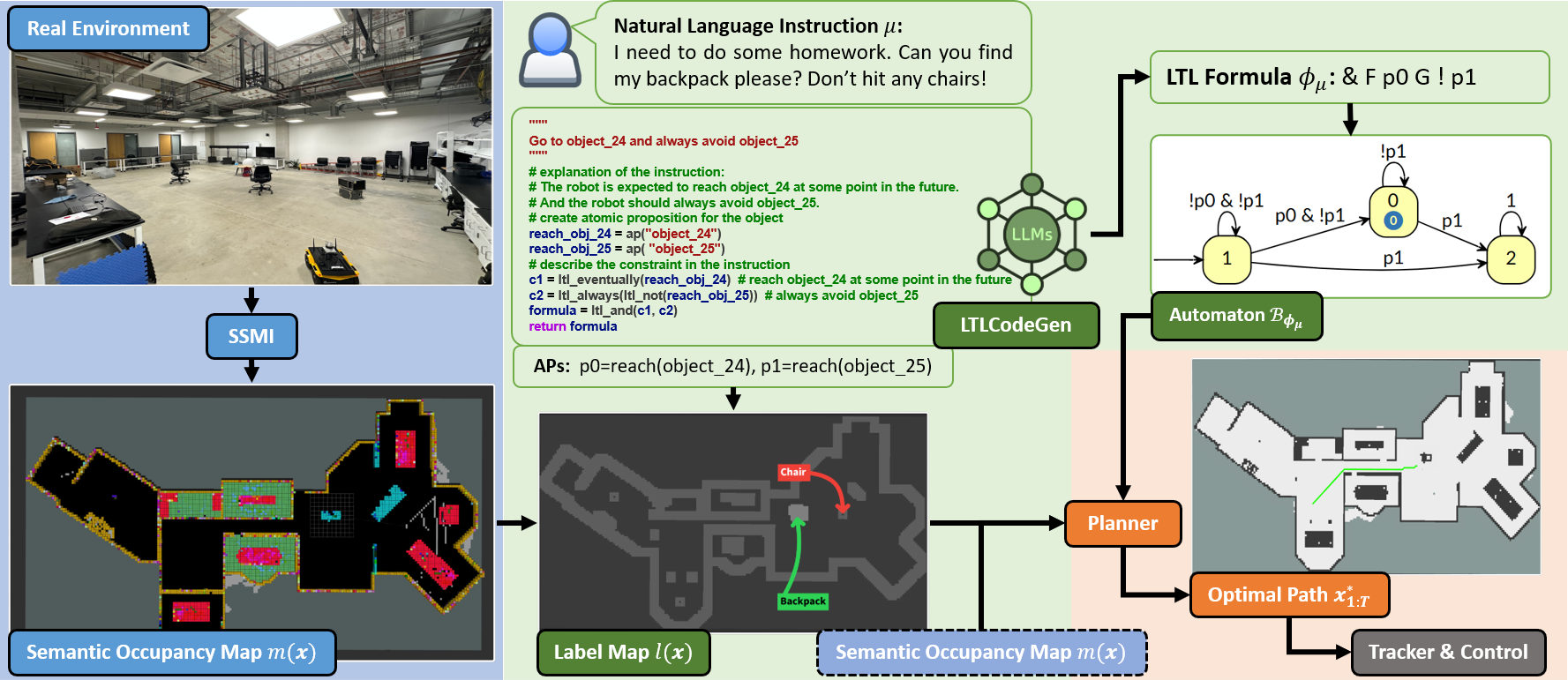}
    \caption{Overview of robot task planning using LTL code generation. Given a semantic occupancy map (left), we convert natural language instructions into a syntactically correct LTL formula using an LLM code generation approach (top). A label map of atomic propositions (bottom, middle), obtained from the map semantic information, and an B{\"u}chi automaton (top, right), obtained from the LTL formula, are used as inputs to a motion planning algorithm to generate a collision-free semantically valid robot path that executes the task.}
    \label{fig:system_architecture}
\end{figure*}

Our approach to the task planning problem described in the previous section begins with constructing a semantic occupancy map, which we discuss in Sec.~\ref{subsec:mapping}. Next, in Sec.~\ref{subsec:ltlcodegen}, we introduce our main contribution, LTLCodeGen, a method to translate natural language to syntactically correct LTL formulas over semantic occupancy map atomic propositions using code generation. Finally, in Sec.~\ref{subsec:planning}, we present a task planning approach that takes the semantic occupancy map and the LTL formula and generates a collision-free robot path that satisfies the task requirements. Fig.~\ref{fig:system_architecture} shows an overview.

\subsection{Semantic Occupancy Mapping}
\label{subsec:mapping}

Before the robot can interpret a desired task, it needs a representation of the environment that captures occupancy (for collision avoidance) and semantic classes (for task grounding). We assume that the robot is able to explore the environment first (either autonomously or through teleoperation) to construct a semantic occupancy map (Definition~\ref{def:semantic_occupancy_map}) using an RGB and range sensor.  Our approach requires three components: an odometry algorithm (e.g., Direct LiDAR Odometry (DLO) \cite{chen2022direct}) that estimates the robot's pose, a semantic segmentation algorithm (e.g., YOLO \cite{yolo2023}) that assigns semantic labels from $\calC$ to image pixels in real time, and a semantic occupancy mapping algorithm (e.g., SSMI \cite{ssmi2023}) to fuse semantically labeled point clouds, constructed from the segmentation and range measurements, into a semantic occupancy map. We chose DLO \cite{chen2022direct} for localization, YOLO \cite{yolo2023} for semantic segmentation, and SSMI \cite{ssmi2023} for semantic occupancy mapping due to their efficiency, robustness, and the availability of open-source code. Other alternatives would also be compatible with our approach.

While SSMI \cite{ssmi2023} constructs a 3D semantic occupancy map, in our experiments we consider a ground wheeled robot navigating on a 2D plane. We project the 3D map $m_{\text{3D}}$ onto this plane and discard voxels outside of a vertical region of interest $\calZ = \{z | z_\text{ground} < z < z_\text{ceiling}\}$. Specifically, we obtain a 2D semantic occupancy map $m$ as follows:
\begin{inparaenum}
   \item $m(x,y) = \textsc{unknown}$ if the entire vertical column of voxels at $(x,y)$ is unobserved;
   \item $m(x,y) = \textsc{free}$ if the vertical column of voxels at $(x,y)$ contains observed voxels and all of them are free;
   \item $m(x,y) = \textsc{null}$ if all the occupied voxels in the column at $(x,y)$ are labeled with $\textsc{null}$;
   \item otherwise, we pick the semantic class from $\bigcup_{z \in \calZ} \{m_\text{3D}(x,y,z)\} \setminus \{\textsc{unknown}, \textsc{free}, \textsc{null}\}$ with the largest $z$ location.
\end{inparaenum}
The robot uses $m(\bfx)$ to interpret the natural language instructions by identifying regions that contain categories of interest and to ensure collision-free task planning and execution. 

\subsection{Translating Natural Language to LTL}\label{subsec:ltlcodegen}

\begin{table}[t]
\caption{\small Grammar for LTL formulas $\phi$ and $\varphi$.}
\scriptsize
\label{tab:ltl_syntax}
\centering
\begin{tabular}{cccccc}
\hline
$p_c$ & (Atomic Proposition) & $\phi \vee \varphi$ & (Or) & $\phi \bfU \varphi$ & (Until) \\
$\neg \phi$ & (Negation) & $\phi \Rightarrow \varphi$ & (Imply) & $\bfF \phi$ & (Eventually) \\
$\phi \wedge \varphi$ & (And) & $\bfX \phi$ & (Next) & $\bfG \phi$ & (Always) \\
\hline
\end{tabular}
\end{table}

This section describes our main contribution, an approach to translate natural language instructions to LTL using code generation, ensuring syntactic correctness of the resulting formulas. LTL is a widely used and sufficiently expressive formalism for expressing a variety of robot tasks \cite{PNUELI198145}. 

LTL formulas are obtained from a set of atomic propositions, logic operators ($\wedge,\vee,\neg$), and temporal operators ($\bfU,\bfF,\bfX,\bfG,\Rightarrow$) with syntax summarized in Table~\ref{tab:ltl_syntax}. We assume that the task $\mu$ can be translated to a syntactically co-safe (sc) LTL formula \cite{kupferman2001model}. Any word that satisfies an sc-LTL formula consists of a finite satisfying prefix followed by any infinite continuation that does not affect the formula's truth value. Thus, sc-LTL formulas allow task satisfaction verification and task planning over finite words.

Although LLMs can translate natural language to LTL formulas using several examples provided with the prompt, LTL formulas generated this way are prone to syntax errors, such as missing parentheses or operands.  Our key idea is to use an LLM to generate (Python) code that describes a natural language instruction $\mu$ using a predefined library of functions, defining the LTL logic and temporal operators. The generated code is then executed to output an LTL formula $\phi_\mu$. Our LTLCodeGen approach inherits the robustness and expressiveness capabilities of LLM code generation and the convenience to verify the LTL syntax as a consequence of the code syntax correctness.

\begin{mycodebox}[label=code:symbol_grounding]{text}{Object Identification}
Your task is to convert the object names to their unique id based on given object id correspondences. Use the object IDs provided in the object ID correspondences for conversion. The conversion should take the context of each sentence into account, so that objects can be correctly correlated to the text.
Here are a few examples:
Object ID correspondence:
    'object_28' : 'refrigerator'
    'object_31' : 'bottle'
    ...
Input text: Take the teddy bear, then pick the bottle. Always avoid the refrigerator.
Output text: Take object_36, then pick object_31. Always avoid the object_28.
...
Using the provided examples, convert the objects in the following text into their unique IDs.
Object ID correspondence: {object ids} 
Input text: {natural language instruction}
Output text: 
\end{mycodebox}

\textbf{Object Category Identification}: 
To ground the semantics of the natural language instruction $\mu$ with respect to the categories in the semantic occupancy map $m(\bfx)$, LTLCodeGen first rephrases the instruction $\mu$ with unique IDs of semantic classes in $\calC$. As shown in Code~\ref{code:symbol_grounding}, we first describe the task of replacing the objects in the instruction with unique IDs. Then, we provide examples of inputs and corresponding expected outputs to the LLM. Finally, we provide the unique IDs of semantic classes in $\calC$, then ask for the instruction $\mu_\calC$ rephrased with IDs, from which we identify categories $\calC_\mu$ present in both the map and the instruction.

\textbf{Code Generation}:
We first briefly describe the code writing task in comments. Then, to make sure that the LLM generates code without undefined variables or functions, we show the available variables and library functions via import statements. As shown in Code~\ref{code:provide_functions}, we first import the function \texttt{ap(obj)} for creating atomic propositions, and then the predefined LTL operators, such as \texttt{ltl\_and(a, b)}. Code~\ref{code:ltl_and} shows the implementation of \texttt{ltl\_and(a, b)}. Other LTL operator functions are defined similarly. The global variable \texttt{prefix} is used to control the LTL format. It is not necessary to provide the implementations of the library functions to the LLM, which wastes the limited number of input tokens. We hide the implementation details of these LTL operator functions and, instead, show examples of how to use them as in Code~\ref{code:provide_examples}.

\begin{mycodebox}[label=code:provide_functions]{python}{Actions and Predefined Functions}
# Please help write code to translate the instruction into an LTL formula.
# Necessary functions and variables are imported.
from ltl_operators import ap  # ap(obj)
from ltl_operators import ltl_and, ltl_or, ltl_not, ltl_until, ltl_eventually, ltl_always, ltl_imply
\end{mycodebox}
\begin{mycodebox}[label=code:ltl_and]{python}{Implementation of \texttt{ltl\_and}}
prefix = True  # The generated LTL is in prefix format
def ltl_and(a: str, b: str):
    if prefix:
        return f"& {a} {b}"
    return f"({a}) & ({b})"
\end{mycodebox}
\begin{mycodebox}[label=code:provide_examples]{python}{Code Example}
def example_1():
    """
    Reach object_2 and object_1
    """
    # explanation of the instruction:
    # object_1 and object_2 are both reached at some point in the future, but no specific order is mentioned explicitly or implicitly.
    # create atomic propositions for objects
    reach_obj_2 = ap("object_2")  # reach(object_2)
    reach_obj_1 = ap("object_1")  # reach(object_1)
    # describe the constraints in the instruction
    c1 = ltl_eventually(reach_obj_2) # Reach object_2...
    c2 = ltl_eventually(reach_obj_1) # Reach object_1...
    c3 = ltl_and(c1, c2)  # Reach object_2 and object_1
    return c3
\end{mycodebox}
The code examples not only show the basic usage of the predefined functions but also demonstrate how to write the code based on the rephrased instruction $\mu_\calC$. The LLM should first explain the instruction in comments, then create atomic propositions for objects involved in the instruction, next describe the instruction with the LTL operator functions, and finally return the LTL formula.

As shown in Code~\ref{code:provide_instruction}, we describe the code writing task again with the requirements for the output format, i.e., the LLM output should only contain the complete implementation of the \texttt{question} function for the instruction $\mu_\calC$.

\begin{mycodebox}[label=code:provide_instruction]{python}{Provide the Instruction to Translate}
# Now, please finish the following Python code for translating the instruction to LTL formula.
# The returned output should only contain the code that starts with `def` and ends with `return` statement.
def question():
    """
    {instruction}{previous_answer}{failure_reason}
    """
\end{mycodebox}

\textbf{Code Syntax Checks}:
The LLM may generate code that contains syntax errors, such as undefined functions, too many arguments for a function call, etc.  To ensure that the generated code is syntactically correct, we execute it and provide the error message from any syntax errors back to the LLM. If an error is detected, we fill the \texttt{previous\_answer} and \texttt{failure\_reason} fields in Code~\ref{code:provide_instruction} with the latest generated code and a short description of the error, respectively. In our practice, GPT-4o~\cite{openai2024gpt4technicalreport} rarely generates Python code with syntax errors, which are easy to fix in a second query to the LLM. When the code is free of syntax errors, it produces an LTL formula $\phi_\mu$ free of LTL syntax errors.

\subsection{Planning}\label{subsec:planning}

Given the atomic propositions $\AP_\mu$ related to the instruction $\mu$ and the LTL formula $\phi_\mu$, synthesized by LTLCodeGen, we can convert $\phi_\mu$ into a B{\"u}chi automaton via a translation tool, such as Spot~\cite{spot}. B{\"u}chi automata are more expressive than LTL formulas, and a B{\"u}chi automaton that recognizes the same language as an LTL formula can always be constructed \cite{wolper1983reasoning}.

\begin{definition}
A deterministic B{\"u}chi automaton is a tuple $\calB=(\calQ, \Sigma, T, \calF, q_1)$, where $\calQ$ is a finite set of states, $\Sigma$ is a finite set of inputs, $T: \calQ \times \Sigma \to \calQ$ is a transition function that specifies the next state $T(q, \sigma)$ from state $q\in\calQ$ and input $\sigma \in \Sigma$, $\calF \subseteq \calQ$ is a set of final (accepting) states, and $q_1 \in \calQ$ is an initial state. 
\end{definition}

Then, to plan a collision-free robot path that satisfies the LTL formula $\phi_\mu$ obtained from LTLCodeGen, we first construct a product planning graph $\calG$ based on the (semantic) occupancy map $m(\bfx)$, the label map $l(\bfx)$, and the automaton $\calB_{\phi_\mu}$. 
With eight-direction movements (cardinal and diagonal) in the (semantic) occupancy map $m(\bfx)$, we construct the graph as $\calG$ $=(\calV, \calE)$, where $\calV$ $=\calX \times \calQ$ is the set of nodes, $\calE$ is the set of transitions from $\bfv_i$$=(\bfx_i, q_i)$ to $\bfv_j$$=(\bfx_j, q_j)$ with a cost $d(\bfx_i,\bfx_j)$. Such a transition exists when
\begin{inparaenum}
    \item $\bfx_j$ can be reached from $\bfx_i$ by a single-step movement in the map $m(\bfx)$;
    \item both $\bfx_i$ and $\bfx_j$ are collision-free: $m(\lambda \bfx_i + (1-\lambda)\bfx_j) = \textsc{free}$, $\forall \lambda \in [-r_o/\alpha,1+r_o/\alpha]$, where $\alpha=\left\|\bfx_i-\bfx_j\right\|_2$, and $r_o$ is a safe margin;
    \item the automaton transitions are respected: $q_j = T(q_i, l(\bfx_i))$.
\end{inparaenum}

Finally, we apply the A* algorithm~\cite{Doran1966ExperimentsWT} to search the graph $\calG$ with a consistent LTL heuristic \cite{dai2024optimal}, defined as:
\begin{equation} \label{eq:ltl_heuristics}
\begin{aligned}
    h(\bfx, q) &= \min_{\bfy \in \calX}\left[ d(\bfx, \bfy) + g(l(\bfy), T(q, l(\bfy))) \right], \\
    h(\bfx, q) &= 0,\quad \forall q \in \calF,
\end{aligned} 
\end{equation}
where the function $g: 2^{\AP} \times \calQ \mapsto \bbR_{\ge0}$ is defined as:
\begin{align}
    c_l(l_1, l_2) &= \min_{\bfx_1, \bfx_2: l(\bfx_1)=l_1, l(\bfx_2)=l_2} c(\bfx_1, \bfx_2), \\
    g(l, q) &= \min_{l' \in 2^{\AP}} c_l(l, l') + g(l', T(q, l')),
\end{align}
which can be pre-computed via dynamic programming on $\calB_{\phi_\mu}$.
The search terminates when the accepting state in $\calF$ is reached, meaning the path satisfies the LTL formula $\phi_\mu$ and, thus, the corresponding natural language instruction $\mu$. If an accepting path $\bfx_{1:T}$ is found, a tracking controller is then used to drive the robot along the path.

\section{Experiments}\label{sec:experiments}

We conduct a series of comprehensive experiments to evaluate LTLCodeGen and our complete task planning method and compare the performance against multiple baselines. First, we compare LTLCodeGen to the LLM-based NL2LTL~\cite{dai2024optimal} and two fine-tuned LLM models~\cite{dataefficientltl2023} on three datasets (Drone, Cleanup, and Pick) to gauge translation accuracy from natural language to LTL. Next, we demonstrate our task planning approach in both simulation and real-world environments, where tasks of varying complexity are tested. We also compare our method to an end-to-end LLM-based task planner on the GameTraversalBenchmark (GTB) \cite{gtb2024}. Finally, we conduct an ablation study to investigate how different LLMs, as well as the inclusion of explanations and comments, affect the performance of LTLCodeGen. 





\begin{table}[t] 
    \centering
    \caption{\small LTL translation accuracy (\%).  Results from the BART baselines are from the original paper \cite{dataefficientltl2023}. BART-FT-RAW-human is shown as BART-human in the table, and BART-FT-RAW-synthetic is shown as BART-syn.}
    \label{tab:comp_ltl_accuracy}
    \resizebox{!}{1cm}{
    \begin{tabular}{l|cc|cc|cc}
    \hline
                    & \multicolumn{2}{c|}{Drone} & \multicolumn{2}{c|}{Cleanup} & \multicolumn{2}{c}{Pick} \\ \hline
                    & human & syn. & human & syn. & human & syn. \\ \hline
        LTLCodeGen & \textbf{99.87} & \textbf{99.94} & \textbf{99.05} & \textbf{98.99} & \textbf{99.19} & \textbf{98.92} \\
        NL2LTL~\cite{dai2024optimal} & 98.66 & 98.30 & 95.36 & 95.21 & 97.31 & 97.18 \\
        BART-human & 90.78 & N/A & 97.84 & N/A & 95.97 & N/A \\
        BART-syn.~\cite{dataefficientltl2023} & 69.39 & N/A & 78.00 & N/A & 81.45 & N/A \\ \hline
    \end{tabular}
    }
\end{table}

\subsection{Evaluation of LTLCodeGen}
Using GPT-4o~\cite{openai2024gpt4technicalreport}, we evaluate our LTLCodeGen and three baselines, NL2LTL~\cite{dai2024optimal}, BART-FT-RAW-human and BART-FT-RAW-synthetic~\cite{dataefficientltl2023} on three datasets (Drone, Cleanup, and Pick) from the fine-tuned LLM \cite{dataefficientltl2023} paper. The three datasets provide both human-written and LLM-augmented synthesized natural language instructions.
For each dataset, we adjust the prompt for LTLCodeGen and NL2LTL based on a few examples randomly drawn from the datasets. The prompt for each dataset is given only 20 (Drone), 12 (Cleanup) and 5 (Pick) examples. Our method and NL2LTL are both tested with all human-written and synthetic instructions. In contrast, BART-FT-RAW-human is evaluated by 5-split cross-validation on human-written data, and BART-FT-RAW-synthetic is trained with synthetic data and tested with human-written data. As shown in Table~\ref{tab:comp_ltl_accuracy}, LTLCodeGen outperforms the baselines significantly on both human-written and synthesized instructions, which indicates that LTLCodeGen generalizes well and generates the correct LTL formulas robustly. NL2LTL also performs better than the fine-tuned LLMs but worse than LTLCodeGen because LLMs are not as good at writing LTL formulas. In our experiments, NL2LTL fails mostly due to the LTL syntax errors that LLMs cannot fix within three retries or due to being semantically mismatched with the instructions.

\subsection{Evaluation on GameTraversalBenchmark (GTB)}
\begin{figure}[t]
    \centering
    \includegraphics[width=\linewidth]{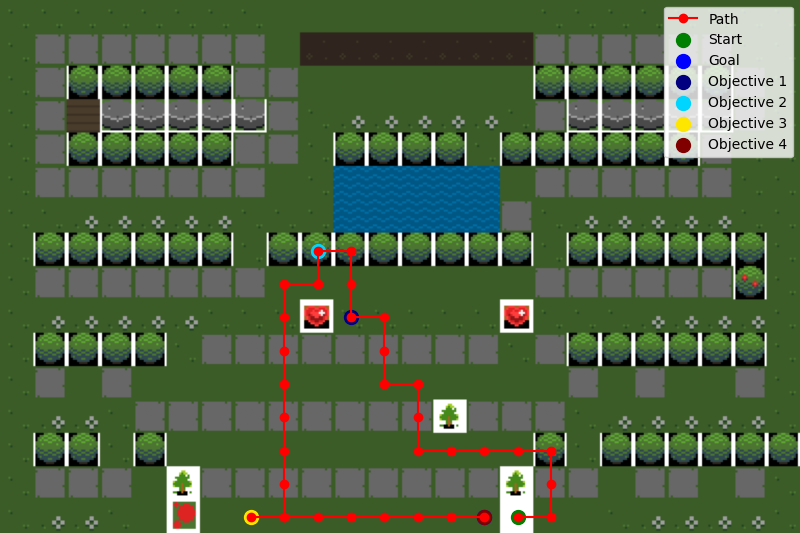}
    \caption{Example path generated by our method for a GTB map. The robot is asked to \textit{\color{BlueViolet}obtain the heartstone}, \textit{\color{Cyan}rally the creatures of Verdanthorn}, then \textit{\color{Yellow}defeat Drakon's Lieutenants}, and finally \textit{\color{Brown}defeat Drakon and restore peace to Verdanthorn}. Note that bush tiles are considered walkable in this map.}
    \label{fig:gtb_example}
\end{figure}
\begin{table}[t]
    \centering
    \caption{\small Evaluation of our task planning method on the GameTraversalBenchmark (GTB) \cite{gtb2024}. Higher Accuracy and lower MPL indicate better performance. Both methods are evaluated using GPT-4o~\cite{openai2024gpt4technicalreport}. Results for GTB \cite{gtb2024} are gathered by using GPT-4o to plan the path directly via prompting as described in their work.}
    \begin{tabular}{l|cc}
        \hline
        \textbf{Method} & \textbf{Accuracy (\%)} $\uparrow$ & \textbf{MPL} $\downarrow$ \\ \hline
        GTB \cite{gtb2024} & $7.84 $ & $85.42 $ \\
        Ours & $\mathbf{78.91}$ & $\mathbf{82.49}$ \\ \hline
    \end{tabular}
    \label{tab:gtb_comparison}
\end{table}

We evaluate our complete task planning approach on the GameTraversalBenchmark (GTB) \cite{gtb2024}. GTB is an LLM-generated dataset that provides 150 different 2D maps with various objects, game stories, and task objectives. For each experiment, we convert the map into the 2D semantic occupancy map format and generate an instruction that combines all the task objectives. For example, if the objectives are \emph{objective\textsubscript{1}}, \emph{objective\textsubscript{2}}, and \emph{objective\textsubscript{3}} the instruction provided to the LLM would be: Complete \emph{objective\textsubscript{1}}, then complete \emph{objective\textsubscript{2}}, and then \emph{objective\textsubscript{3}}. Our system consumes the map and the instruction to generate a path that satisfies all the task objectives. We report two metrics in Table~\ref{tab:gtb_comparison}:
\begin{itemize}
    \item \textbf{Accuracy ($\uparrow$):} the percentage of tests where the agent successfully reaches the exact objective coordinates;
    
    \item \textbf{Mean Path Length (MPL) ($\downarrow$):} the average length of paths taken by the agent to complete each task.
\end{itemize}

The results in Fig.~\ref{fig:gtb_example} and Table~\ref{tab:gtb_comparison} show that our system outperforms the end-to-end GPT-4o-based planner from GTB~\cite{gtb2024}. However, there are cases where our planner is unable to generate a path due to the absence of a valid path satisfying the LTL in the environment. For example, achieving \emph{objective\textsubscript{1}} may require passing through a later objective, such as \emph{objective\textsubscript{3}}, which conflicts with the generated LTL.

\begin{table*}[t]
    \centering
    \caption{\small LTL generation comparison: LTLCodeGen vs NL2LTL \cite{dai2024optimal}. Two LLMs, GPT-4o and GPT-4o-mini, are tested.}
    \begin{tabular}{l|cccc|cccc|cccc|c}
        Method & \multicolumn{4}{c|}{Success (\%)$\uparrow$} & \multicolumn{4}{c|}{Semantic Failure (\%)$\downarrow$} & \multicolumn{4}{c|}{Syntactic Failure (\%)$\downarrow$} & \multicolumn{1}{c}{Runtime (s)$\downarrow$} \\ \hline
        Tier & 1 & 2 & 3 & 4 & 1 & 2 & 3 & 4 & 1 & 2 & 3 & 4 & Total \\ \hline
        LTLCodeGen-4o & \textbf{100} & \textbf{100} & \textbf{100} & \textbf{70.0} & \textbf{0} & \textbf{0} & \textbf{0} & \textbf{30.0} & \textbf{0} & \textbf{0} & \textbf{0} & \textbf{0} & ${7.07 \pm 4.31}$ \\
        NL2LTL-4o\cite{dai2024optimal} & 100 & 90.0 & 46.7 & 60.0 & 0 & 10.0 & 40.0 & 40.0 & 0 & 0 & 13.3 & 0 & $\mathbf{2.01 \pm 1.12}$
        \\
        LTLCodeGen-4o-NoExplanation & 100 & 100 & 60.0 & 50.0 & 0 & 0 & 40.0 & 50.0 & 0 & 0 & 0 & 0 & ${5.55 \pm 3.20}$ \\
        LTLCodeGen-4o-NoComment & 100 & 100 & 93.3 & 70 & 0 & 0 & 6.7 & 30.0 & 0 & 0 & 0 & 0 & ${6.06 \pm 3.06}$ \\
        \rowcolor{lightgray}
        LTLCodeGen-4o-mini & 90.0 & 70.0 & 26.7 & 0 & 10.0 & 30.0 & 73.3 & 100 & 0 & 0 & 0 & 0 & ${4.90 \pm 3.40}$ \\
        \rowcolor{lightgray}
        NL2LTL-4o-mini & 80.0 & 70.0 & 13.3 & 0 & 20.0 & 10.0 & 26.7 & 80.0 & 0 & 20.0 & 60.0 & 20.0 & ${2.15 \pm 0.93}$ \\
        \rowcolor{lightgray}
        LTLCodeGen-4o-mini-NoExplanation & 100 & 80.0 & 20.0 & 0 & 0 & 20.0 & 80.0 & 100.0 & 0 & 0 & 0 & 0 & ${4.49 \pm 2.22}$ \\
        \rowcolor{lightgray}
        LTLCodeGen-4o-mini-NoComment & 90.0 & 80.0 & 40.0 & 0 & 10.0 & 20.0 & 60.0 & 100.0 & 0 & 0 & 0 & 0 & ${3.77 \pm 1.25}$ 
    \end{tabular}
    \label{tab:LTLCodeGen_vs_nl2ltl}
    \vspace{-1em}
\end{table*}

\subsection{Real-World Experiments}
We evaluate our system in a mock indoor environment containing eight categories: personal items (backpack, laptop, umbrella), common household objects (TV, potted plant, microwave, chairs), and a person for a realistic setting.

The task design includes four tiers of natural language tasks progressively increasing in difficulty:
\begin{itemize}
    \item \textit{Tier 1: Single-object tasks.} 
    Straightforward instructions referencing one object (e.g., ``Visit the chair'').
    
    \item \textit{Tier 2: Multi-object tasks with avoidance.} 
    Missions involving one or more objects while \emph{avoiding} others (e.g., ``Go to the table and avoid chairs'') or \emph{applying a condition} (e.g., ``Go to the backpack if you are near a laptop'').
    
    \item \textit{Tier 3: Multi-object tasks with temporal constraints.} 
    Complex missions requiring multiple visits or precise ordering (e.g., ``Visit the desk, then the couch, and do not go by the chair'').
    
    \item \textit{Tier 4: Ambiguous, context-based tasks.} 
    Instructions relying on inference and context rather than explicit object naming (e.g., ``It's valentine's day and I have no date. Let's watch a romcom.'').
\end{itemize}
The evaluation includes 10 scenarios for Tier 1, 10 for Tier 2, 15 for Tier 3, and 10 for Tier 4, totaling 45 scenarios. Our planner runs using LTL specifications generated by LTLCodeGen or NL2LTL \cite{dai2024optimal}, both with GPT-4o and GPT-4o-mini~\cite{openai2024gpt4technicalreport}. The robot's trajectory undergoes visual inspection to determine task success and the following metrics:

\begin{itemize}
    \item \textbf{Success Rate}: 
    Fraction of tasks the robot completes.
    
    \item \textbf{Semantic Error Rate}: 
    Fraction of tasks where the plan does not match user intent.
    
    \item \textbf{Syntactic Error Rate}: 
    Fraction of tasks where the LLM generates invalid LTL after three retries.
    
    \item \textbf{LLM Runtime}: 
    Time required to produce LTL formula.
\end{itemize}

Table~\ref{tab:LTLCodeGen_vs_nl2ltl} shows that LTLCodeGen consistently outperforms NL2LTL. Notably, LTLCodeGen incurs no syntactic errors, thanks to its predefined LTL operator functions. Since LLMs are better at writing code than LTL-specific syntax, this approach delivers stronger correctness guarantees.
Moreover, LTLCodeGen excels in sequential and revisiting tasks, where strict ordering challenges the baseline. Expressing constraints in code appears more manageable for LLMs. 

However, LTLCodeGen requires longer inference times since it generates complete executable code, including comments and explanations, while NL2LTL produces only a single-line LTL formula. Our ablation study (Section~\ref{sec:ablation}) confirms that removing explanations or comments reduces generation length and accelerates inference. Our method sometimes struggles to generate semantically correct results when ambiguity exists in the instruction (Tier 4). Although including explanations and comments in the code helps to resolve most ambiguity in the instructions, how to interpret the semantic constraint robustly remains an open question.

\subsection{Ablation Study}
\label{sec:ablation}
We examine three variations of our method on the same 45 real-world tasks: (1) the full version, (2) a version without task explanations, and (3) a version without line comments in the code. Each variant uses both GPT-4o and GPT-4o-mini \cite{openai2024gpt4technicalreport} to evaluate the impact of model size.

\subsubsection{No Explanation}
As shown in Table~\ref{tab:LTLCodeGen_vs_nl2ltl}, removing task explanations significantly reduces Tier 3 performance and noticeably harms Tier 4. While Tiers 1 and 2 remain relatively straightforward, the revisiting and sequencing in Tier 3 become error-prone without explanations, and Tier 4's contextual ambiguity exacerbates this issue. These findings demonstrate that providing explanations yields a clear net benefit across both GPT-4o and GPT-4o-mini.

\subsubsection{No Comment}
Eliminating line comments has minimal impact on overall performance, likely because LLMs rely more on task explanations than on code annotations. However, GPT-4o shows a slight performance drop in Tier 3, suggesting that comments can clarify complex temporal logic. In contrast, GPT-4o-mini sometimes improves when comments are removed, indicating that extraneous text may add noise for smaller models.

\subsubsection{Model Size and Performance}
Switching from GPT-4o to GPT-4o-mini results in a performance decline, especially for Tier 4 tasks, reinforcing that smaller models struggle more with ambiguity. Tier 3 performance also declines for all variations, reflecting the complexity of sequencing constraints. However, Tiers 1 and 2 experience only modest drops, suggesting that larger models are generally more robust across all difficulty levels.







\section{Conclusion}\label{sec:conclusion}

We introduced an LLM-based planning approach that translates natural language instructions into syntactically correct LTL specifications, using a semantic occupancy map and a motion planning algorithm to generate collision-free paths satisfying the instruction. Utilizing the proficiency of LLMs in code generation is central for our approach to generate syntactically correct LTL. Simulation and real-world experiments show that our LTLCodeGen approach outperforms unstructured LTL synthesis methods and plays a key role in enabling a mobile robot to robustly execute natural language specifications. Our results suggest that LLMs should be utilized according to their strengths deriving from the training data content as opposed to novel concept prompting with limited in-context training.

\bibliographystyle{cls/IEEEtran}
\bibliography{bib/bibliography}

\begin{thebibliography}{10}
\providecommand{\url}[1]{#1}
\csname url@rmstyle\endcsname
\providecommand{\newblock}{\relax}
\providecommand{\bibinfo}[2]{#2}
\providecommand\BIBentrySTDinterwordspacing{\spaceskip=0pt\relax}
\providecommand\BIBentryALTinterwordstretchfactor{4}
\providecommand\BIBentryALTinterwordspacing{\spaceskip=\fontdimen2\font plus
\BIBentryALTinterwordstretchfactor\fontdimen3\font minus \fontdimen4\font\relax}
\providecommand\BIBforeignlanguage[2]{{%
\expandafter\ifx\csname l@#1\endcsname\relax
\typeout{** WARNING: IEEEtran.bst: No hyphenation pattern has been}%
\typeout{** loaded for the language `#1'. Using the pattern for}%
\typeout{** the default language instead.}%
\else
\language=\csname l@#1\endcsname
\fi
#2}}

\bibitem{openai2024gpt4technicalreport}
OpenAI, ``{GPT}-4 {T}echnical {R}eport,'' \emph{arXiv preprint: arXiv 2303.08774}, 2024.

\bibitem{llama22023}
H.~Touvron, L.~Martin, \emph{et~al.}, ``{Llama 2: Open Foundation and Fine-Tuned Chat Models},'' \emph{arXiv preprint: arXiv 2307.09288}, 2023.

\bibitem{deepseek2025}
DeepSeek-AI, ``{DeepSeek-R1: Incentivizing Reasoning Capability in LLMs via Reinforcement Learning},'' \emph{arXiv preprint: arXiv 2501.12948}, 2025.

\bibitem{saycan2023}
B.~Ichter, A.~Brohan, \emph{et~al.}, ``{Do As I Can, Not As I Say: Grounding Language in Robotic Affordances},'' in \emph{Conference on Robot Learning}, 2023.

\bibitem{roco2024}
Z.~Mandi, S.~Jain, and S.~Song, ``{RoCo: Dialectic Multi-Robot Collaboration with Large Language Models},'' in \emph{IEEE International Conference on Robotics and Automation}, 2024.

\bibitem{gtb2024}
M.~U. Nasir, S.~James, and J.~Togelius, ``{GameTraversalBenchmark: Evaluating Planning Abilities Of Large Language Models Through Traversing 2D Game Maps},'' \emph{Advances in Neural Information Processing Systems}, 2024.

\bibitem{dai2024optimal}
Z.~Dai, A.~Asgharivaskasi, T.~Duong, S.~Lin, M.-E. Tzes, G.~Pappas, and N.~Atanasov, ``{Optimal Scene Graph Planning with Large Language Model Guidance},'' in \emph{IEEE International Conference on Robotics and Automation}, 2024.

\bibitem{codepolicy2023}
J.~Liang, W.~Huang, F.~Xia, P.~Xu, K.~Hausman, B.~Ichter, P.~Florence, and A.~Zeng, ``{Code as Policies: Language Model Programs for Embodied Control},'' in \emph{IEEE International Conference on Robotics and Automation}, 2023.

\bibitem{chatenv2023}
X.~Zhao, M.~Li, C.~Weber, M.~B. Hafez, and S.~Wermter, ``{Chat with the Environment: Interactive Multimodal Perception Using Large Language Models},'' in \emph{IEEE/RSJ International Conference on Intelligent Robots and Systems}, 2023.

\bibitem{progprompt2023}
I.~Singh, V.~Blukis, A.~Mousavian, A.~Goyal, D.~Xu, J.~Tremblay, D.~Fox, J.~Thomason, and A.~Garg, ``{P}rog{P}rompt: {G}enerating {S}ituated {R}obot {T}ask {P}lans using {L}arge {L}anguage {M}odels,'' in \emph{IEEE International Conference on Robotics and Automation}, 2023.

\bibitem{smartllm2024}
S.~S. Kannan, V.~L.~N. Venkatesh, and B.-C. Min, ``{SMART-LLM: Smart Multi-Agent Robot Task Planning using Large Language Models},'' \emph{arXiv preprint: arXiv 2309.10062}, 2024.

\bibitem{spine2024}
Z.~Ravichandran, V.~Murali, M.~Tzes, G.~J. Pappas, and V.~Kumar, ``{SPINE: Online Semantic Planning for Missions with Incomplete Natural Language Specifications in Unstructured Environments},'' \emph{arXiv preprint: arXiv 2410.03035}, 2024.

\bibitem{pddlllm2024}
T.~Silver, S.~Dan, K.~Srinivas, J.~B. Tenenbaum, L.~Kaelbling, and M.~Katz, ``{Generalized Planning in PDDL Domains with Pretrained Large Language Models},'' in \emph{AAAI/IAAI/EAAI}, 2024.

\bibitem{symbolgrounding2011}
S.~Tellex, T.~Kollar, S.~Dickerson, M.~R. Walter, A.~G. Banerjee, S.~Teller, and N.~Roy, ``{Approaching the Symbol Grounding Problem with Probabilistic Graphical Models},'' \emph{AI Magazine}, 2011.

\bibitem{demo2code2023}
H.~Wang, G.~Gonzalez-Pumariega, Y.~Sharma, and S.~Choudhury, ``{Demo2Code: From Summarizing Demonstrations to Synthesizing Code via Extended Chain-of-Thought},'' \emph{arXiv preprint: arXiv 2305.16744}, 2023.

\bibitem{kambhampati2024llmscantplanhelp}
S.~Kambhampati, K.~Valmeekam, L.~Guan, M.~Verma, K.~Stechly, S.~Bhambri, L.~Saldyt, and A.~Murthy, ``{LLMs Can't Plan, But Can Help Planning in LLM-Modulo Frameworks},'' \emph{arXiv preprint: arXiv 2402.01817}, 2024.

\bibitem{valmeekam2023planningabilitieslargelanguage}
K.~Valmeekam, M.~Marquez, S.~Sreedharan, and S.~Kambhampati, ``{On the Planning Abilities of Large Language Models : A Critical Investigation},'' \emph{arXiv preprint: arXiv 2305.15771}, 2023.

\bibitem{innermonologue2022}
W.~Huang, F.~Xia, \emph{et~al.}, ``{Inner Monologue: Embodied Reasoning through Planning with Language Models},'' \emph{arXiv preprint: arXiv 2207.05608}, 2022.

\bibitem{instruct2act2023}
S.~Huang, Z.~Jiang, H.~Dong, Y.~Qiao, P.~Gao, and H.~Li, ``{Instruct2Act: Mapping Multi-modality Instructions to Robotic Actions with Large Language Model},'' \emph{arXiv preprint: arXiv 2305.11176}, 2023.

\bibitem{ssmi2023}
A.~Asgharivaskasi and N.~Atanasov, ``{Semantic Octree Mapping and {S}hannon Mutual Information Computation for Robot Exploration},'' \emph{IEEE Transactions on Robotics}, 2023.

\bibitem{yolo2023}
\BIBentryALTinterwordspacing
G.~Jocher, J.~Qiu, and A.~Chaurasia, ``{Ultralytics YOLO},'' Jan. 2023. [Online]. Available: \url{https://github.com/ultralytics/ultralytics}
\BIBentrySTDinterwordspacing

\bibitem{dataefficientltl2023}
J.~Pan, G.~Chou, and D.~Berenson, ``{Data-Efficient Learning of Natural Language to Linear Temporal Logic Translators for Robot Task Specification},'' in \emph{IEEE International Conference on Robotics and Automation}, 2023.

\bibitem{climbing2020}
E.~M. Bender and A.~Koller, ``{Climbing towards {NLU}: {On} Meaning, Form, and Understanding in the Age of Data},'' in \emph{Annual Meeting of the Association for Computational Linguistics}, 2020.

\bibitem{gubelmann2024}
R.~Gubelmann, ``{Pragmatic Norms Are All You Need {--} Why The Symbol Grounding Problem Does Not Apply to {LLM}s},'' in \emph{Conference on Empirical Methods in Natural Language Processing}, Y.~Al-Onaizan, M.~Bansal, and Y.-N. Chen, Eds.\hskip 1em plus 0.5em minus 0.4em\relax Association for Computational Linguistics, 2024.

\bibitem{autotamp2024}
Y.~Chen, J.~Arkin, C.~Dawson, Y.~Zhang, N.~Roy, and C.~Fan, ``{AutoTAMP: Autoregressive Task and Motion Planning with LLMs as Translators and Checkers},'' in \emph{IEEE International Conference on Robotics and Automation}, 2024.

\bibitem{nl2tl2024}
Y.~Chen, R.~Gandhi, Y.~Zhang, and C.~Fan, ``{NL2TL: Transforming Natural Languages to Temporal Logics using Large Language Models},'' \emph{arXiv preprint: arXiv 2305.07766}, 2024.

\bibitem{tulip2011}
T.~Wongpiromsarn, U.~Topcu, N.~Ozay, H.~Xu, and R.~M. Murray, ``{TuLiP: a Software Toolbox for Receding Horizon Temporal Logic Planning},'' in \emph{International Conference on Hybrid Systems: Computation and Control}, 2011.

\bibitem{vasile2013sampling}
C.~I. Vasile and C.~Belta, ``{Sampling-based Temporal Logic Path Planning},'' in \emph{IEEE/RSJ International Conference on Intelligent Robots and Systems}, 2013.

\bibitem{optimaltl2016}
J.~Fu, N.~Atanasov, U.~Topcu, and G.~J. Pappas, ``{Optimal Temporal Logic Planning in Probabilistic Semantic Maps},'' in \emph{IEEE International Conference on Robotics and Automation}, 2016.

\bibitem{armeni20193d}
I.~Armeni, Z.-Y. He, J.~Gwak, A.~R. Zamir, M.~Fischer, J.~Malik, and S.~Savarese, ``{3D Scene Graph: A Structure for Unified Semantics, 3D Space, and Camera},'' in \emph{IEEE International Conference on Computer Vision}, 2019.

\bibitem{rosinol20203d}
A.~Rosinol, A.~Gupta, M.~Abate, J.~Shi, and L.~Carlone, ``{3D Dynamic Scene Graphs: Actionable Spatial Perception with Places, Objects, and Humans},'' in \emph{Robotics: Science and Systems}, 2020.

\bibitem{kimera}
A.~Rosinol, A.~Violette, M.~Abate, N.~Hughes, Y.~Chang, J.~Shi, A.~Gupta, and L.~Carlone, ``{Kimera: From SLAM to Spatial Perception with 3D Dynamic Scene Graphs},'' \emph{The International Journal of Robotics Research}, 2021.

\bibitem{hughes2022hydra}
N.~Hughes, Y.~Chang, and L.~Carlone, ``{H}ydra: {A} {R}eal-time {S}patial {P}erception {S}ystem for {3D} {S}cene {G}raph {C}onstruction and {O}ptimization,'' in \emph{Robotics: Science and Systems}, 2022.

\bibitem{amiri2022reasoning}
S.~Amiri, K.~Chandan, and S.~Zhang, ``{R}easoning with {S}cene {G}raphs for {R}obot {P}lanning under {P}artial {O}bservability,'' \emph{IEEE Robotics and Automation Letters (RAL)}, 2022.

\bibitem{socratic2022}
A.~Zeng, M.~Attarian, \emph{et~al.}, ``{S}ocratic {M}odels: {C}omposing {Z}ero-{S}hot {M}ultimodal {R}easoning with {L}anguage,'' \emph{arXiv preprint: arXiv 2204.00598}, 2022.

\bibitem{chen2022direct}
K.~Chen, B.~T. Lopez, A.-a. Agha-mohammadi, and A.~Mehta, ``{D}irect {L}i{D}{A}{R} {O}dometry: {F}ast {L}ocalization {W}ith {D}ense {P}oint {C}louds,'' \emph{IEEE Robotics and Automation Letters}, 2022.

\bibitem{PNUELI198145}
A.~Pnueli, ``{T}he {T}emporal {S}emantics of {C}oncurrent {P}rograms,'' \emph{Theoretical Computer Science}, 1981.

\bibitem{kupferman2001model}
O.~Kupferman and M.~Y. Vardi, ``{M}odel {C}hecking of {S}afety {P}roperties,'' \emph{Formal Methods in System Design}, 2001.

\bibitem{spot}
A.~Duret-Lutz, E.~Renault, \emph{et~al.}, ``From {S}pot 2.0 to {S}pot 2.10: {W}hat's {N}ew?'' in \emph{International Conference on Computer Aided Verification (CAV)}, 2022.

\bibitem{wolper1983reasoning}
P.~Wolper, M.~Y. Vardi, and A.~P. Sistla, ``Reasoning about infinite computation paths,'' in \emph{Annual Symposium on Foundations of Computer Science}, 1983.

\bibitem{Doran1966ExperimentsWT}
J.~Doran and D.~Michie, ``{Experiments with the Graph Traverser Program},'' \emph{Proceedings of the Royal Society of London. Series A. Mathematical and Physical Sciences}, 1966.

\end{thebibliography}

\end{document}